# Gazing into the Abyss: Real-time Gaze Estimation

George He
Stanford University
georgehe@stanford.edu

Sami Oueida
Stanford University
soueida@stanford.edu

Tucker Ward
Stanford University
tlward@stanford.edu

## Abstract

*Gaze and face tracking algorithms have traditionally battled a compromise between computational complexity and accuracy; the most accurate neural net algorithms cannot be implemented in real time, but less complex real-time algorithms suffer from higher error. This project seeks to better bridge that gap by improving on real-time eye and facial recognition algorithms in order to develop accurate, real-time gaze estimation with an emphasis on minimizing training data and computational complexity. Our goal is to use eye and facial recognition techniques to enable users to perform limited tasks based on gaze and facial input using only a standard, low-quality web cam found in most modern laptops and smart phones and the limited computational power and training data typical of those scenarios. We therefore identified seven promising, fundamentally different algorithms based on different user features and developed one outstanding, one workable, and one honorable mention gaze tracking pipelines that match the performance of modern gaze trackers while using no training data.*

## 1. Introduction

Eye and face tracking has innumerable applications in human-computer interaction, medicine, computer vision, and other fields. Detection and estimation of gaze, an observable indicator for human attention and focus, has been studied and recorded throughout the past two centuries. [9].

Many gaze and face tracking techniques have been developed, but the most accurate have poor user experiences. On one extreme, scleral search coils are highly accurate but are in direct contact with the the eye [5], while other approaches rely on dataset heavy and computationally intensive neural nets that provide limited insight into the features and underlying training weights [20]. Furthermore, these neural net-based approaches, while accurate, are far too computationally difficult to implement in real time on a standard laptop or mobile device.

On the other hand, many algorithms exist to estimate gaze and face tracking based on relatively simple features, from skin color [12] to more mathematical approaches like longest line scanning [13]. These algorithms are simple enough to run in real time on mobile devices, but the accuracy is typically only sufficient to provide coarse input.

We believe that gaze estimation and facial recognition can be implemented without either set of undesirable drawbacks and still produce highly accurate results. Our approach is to combine several real-time algorithms that rely on different characteristics to improve accuracy to the point where gaze and face estimation can be used for fine-grain control of a computational device. For example, we may enable the user to pan a monitor and rotate in accordance with user face or eye movement. We limit ourselves to a tiny training set (only a few images) and poor quality cameras to accurately reflect the conditions most users encounter daily. By combining different geometric, color, and mathematical models, we hope to improve gaze estimation to nearly the accuracy of a neural net, but in real time on a mobile device.

We chose seven fundamental techniques to contribute features, including skin tone [12], longest line scanning [13], occluded circular eye matching (OCEM) [13], Hough circle detection [6], "Eigenfaces" [14], as well as a number of other geometric techniques such as Harris edge detection. While not all of these techniques proved useful, we implemented each and experimented to find an optimal mix that create an accurate real-time gaze detector and several techniques that come close to real-time.

There are a number of key challenges to our approach, including some inherent to gaze and face tracking. First, human gaze and face orientation are inherently very noisy; users subconsciously and consciously move their eyes and head all the time, and this noise must be separated from bona fide commands the user issues. Second, our target users are average people with the low-quality cameras and modest computational power found in most laptops and mobile devices today. Third, most devices will be used by just one user, and therefore the training data for any given user or device is extremely limited.

## 2. Background and Related Work

We cannot hope to do cover all relevant work, but will provide a summary of the gaze tracking field. Numerous techniques exist to determine head orientation and gaze tracking. In broad strokes, tracking techniques are either intrusive or remote. Intrusive techniques rely on physical contact (e.g., using specialized contact lenses) to track a user's gaze. We focus on remote techniques.

Remote techniques can be generally broken into two categories: appearance-based and geometry-based. Appearance-based features rely on color or other visual characteristics of the subject to determine face and eye position, while geometric features use geometric proper-ties such as occluded circular eye matching or longest line scanning to identify and model the pupils, eyes, etc. and make a best estimate of gaze location.

Numerous researchers have used appearance and geometric-based techniques to great success. Some of the simplest, and often remarkably accurate, techniques include skin tone detection [12], longest line scanning [13], and edge detection to localize the face and eyes, and predict gaze based on eye position. These techniques typically rely on the assumption that the key is a sphere, and that gaze can be predicted as the projection of the normal from the center of the pupil. The advantage to these techniques is that they can rely on relatively simple facial and eye detection/recognition techniques, but their accuracy is somewhat bounded by the underlying assumptions they require. More advanced versions of these techniques can be calibrated to an individual to greatly improve accuracy [8].

Other researchers pioneered highly intricate methods. Specialized hardware can be used to detect Purkinje images [19] (specific reflections off different parts of the eye) that can localize the pupil and eye orientation with extreme accuracy, but the hardware is expensive [8]. State-of-the-art convolutional neural network techniques produce field-leading accuracy of over 80% using only off-the-shelf webcams, but even on a high-end desktop from 2016 the technique could manage only about 20 frames per second [1]. Neural networks are far more accurate than any other technique, but it will be 7-10 years before standard mobile devices are sufficiently powerful to use them [1].

One final and important dimension of existing research is boosting the detectability of the face, eye, and pupil, effectively reducing the complexity of algorithm needed for a given accuracy threshold. The most widely used technique is projecting small amount of infrared light that reflects off the cornea, causing just the pupil to show as an unambiguous white circle on an IR camera. This technique and has been reported to reduce error in gaze tracking estimation by as much as 75% [3]. In most cases, IR boosting requires specialized hardware, but the technique may become common as IR cameras become more ubiquitous as anti-spoofing tools for facial recognition. We suspect that this technique would decrease our error substantially.

We see an opportunity to approach neural network accuracy using combinations of simpler techniques usable today.

## 3. Experimental Approach and Techniques

### 3.1. Experimental Overview

#### 3.1.1 Techniques

Our basic goal is to create a unified pipeline of computationally simple gaze detection techniques that can accurately determine gaze location in real time with minimal or no training data. We chose to implement 7 techniques:

- Longest line scanning (LLS) [13]
- Occluded circular eye matching (OCEM) [13]
- Hough circle detection [6]
- Skin color model [12]
- Eigenfaces model [14]
- Harris/canny edge detector, Gaussian smoothing [4]
- Haar cascades [17]

We then tested combinations of these techniques to create end-to-end pipelines for gaze detection. These techniques rely on different features and have different detection capabilities (eg, edge v face); our goal was to implement a range of techniques with which we could experiment.

It is important to note that we implemented every technique ourselves from scratch, with the exception of Haar cascades and hough transform. As each of these techniques was originally published individually as a paper, we effectively implemented 5 past research papers. In some cases we later substituted opencv equivalents either to provide a baseline or improve performance in the final product. A link to code is provided in the supplementary materials section.

#### 3.1.2 Pipeline Approach

Each technique listed above has its own advantages and disadvantages as discussed below. Each also fills a different role, for example edge or face detection. Some techniques, such as skin color, were implemented in the hope of quickly localizing the face/eyes and then passing this information to a different, faster technique.

We strategically combined our raw detection techniques to create full pipelines to extract key information such as the location of the pupils and corners of the eyes in order to estimate gaze. For example, we experimented with using the skin color model to detect faces and eyes, and then edge detection on the eyes to find pupils and estimate orientation. In another case, we used LLS and OCEM to geometrically detect pupils and estimate orientation of the pupil center. Results are discussed in section 4.

### 3.2. Experimental Techniques

#### 3.2.1 Longest Line Scanning (LLS)

Longest line scanning finds the pupil in two steps. First, it locates the iris by fitting a circle to the iris using least squares estimation. The limbus, or boundary of the iris, should be easily found as the color difference between the white sclera and darker iris. This is relatively well detected by standard edge detection methods. Second, after finding the iris using circle fitting, one fits the longest horizontal line possible within it. As a circle is a suitable enough estimation of an ellipse we can use the property of ellipses that state the center of an ellipse is the midpoint of the longest line connecting two points along the ellipse. So the location of this longest line will run through the middle of the iris, and the midpoint will be a good approximation for the retina. We can then use the retina's location in one of our pipelines to calibrate a retina-to-screen relationship and estimate the gaze, as is discussed more in results.

While LLS estimation does give us a good approximate location of the pupil, there are several key limitations [13]:
- Coverage of parts of the limbus by the eyelids
- Poor quality of the images
- Excessive coverage of the eyes by eyelids Each of these challenges reduces LLS accuracy. Figure 1 shows the results of our LLS implementation. One can see that there are many points of interest, and thus the final pupil location is selected with occluded circular eye matching.

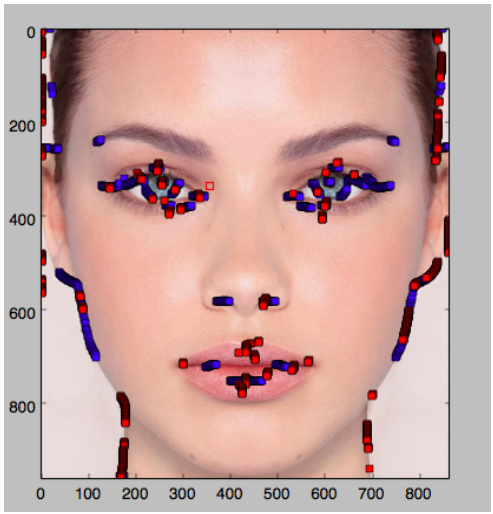

Figure 1. Example of LLS and areas of interest.

#### 3.2.2 Occluded Circular Eye Matching (OCEM)

OCEM selects among the possible iris locations and picks the most likely as the result. In general, OCEM works by constructing an estimated limbus around a test iris point, and then matching that estimate against the image, scoring it based on its fit to the image. Since the limbus is often obstructed in the image we pick a portion, usually the sides, of the limbus to match to the image. The iris location with the highest scored corresponding limbus match is chosen as the true iris location. While this general technique describes OCEM, the results were often too noisy or inaccurate to use in gaze estimation. We thus adopted a refined model in order to better select candidate irises.

#### 3.2.3 Refined LLS with OCEM

The first major improvement was robust eye detection, which greatly reduces the number of inaccurate LLS candidates–LLS tends to find many candidates along any significant color gradient, as you can see in Figure 1. We tried a number of smoothing and filtration techniques to reduce noise and filter poor eye candidates; of those, Gaussian smoothing and Sobel-Fieldman filtration to approximate the gradient of the pixel intensity in the picture worked best. We also considered Roberts Cross, Prewitt, and Scharr filtration, but they did not perform as well.

Second, after determining where the eye is, we create a bounding box for a more thorough search. We then used a windowing technique that includes non-maximal suppression and convolution only in one direction to specifically highlight vertical lines (horizontal gradients), as the limbus has approximately vertical edges around the iris. We found that thresholding the gradient to reduce noise helps increase the accuracy substantially, as the limbus and iris are typically very large features. From that result we then approximate the horizontal location of the iris as the midpoint between the sides of the prospective limbus. This result, in addition to our horizontal bounds from our refined eye window intensity gradient search, gives us the vertical location of the iris, reducing our pupil search to only one dimension. Once we have this result employ LLS (as described above) as a secondary check for iris estimation, and typically the combination of these techniques proved a very powerful estimate of true pupil location, as seen in Figure 2.

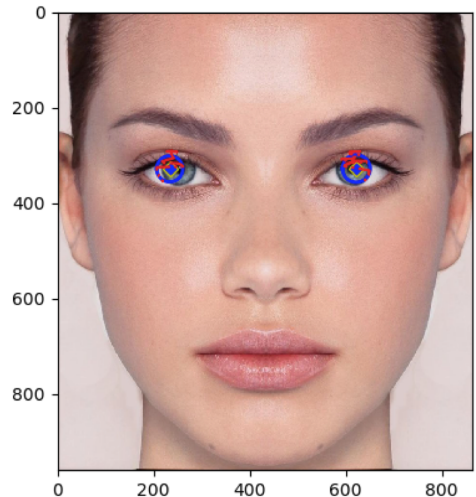

Figure 2. Example of modified OCEM

#### 3.2.4 Hough Circle detection

Hough circle detection can be applied to find pupils [6], and we implemented this technique because of its ability to find uniquely pupil-like features in the eye and estimate accurately the pupil center. Because application of the hough transform can be computationally intensive for larger regions and is prone to mis-classification of circular objects, we only apply the hough circle transform after localizing the general location of the eye. After finding the eye using one of our facial/eye detection techniques (eg, skin color), we apply a smoothing Gaussian layer before running a canny edge detector. This results in a rough outline of important features within the eye. After applying another smoothing Gaussian layer resolve the rigidity resulting from the canny edge filter, we use a Hough circle transform that allows us to find areas within the eye that are likely to correspond with the pupil. Adding a constraint that the resulting circles must be a certain radius, obtained as a function of the size of the detected face, we localize the iris with high accuracy and simply take the center of the computed circle as the pupil's center. The Hough transform is robust against image noise, and detects the iris even if only a part of the outline of the iris appears. This approach solves the common problem that eyelids typically occlude the top and bottom portions of the iris. The result is a highly accurate localization of the iris or pupil, as shown in Figure 3.

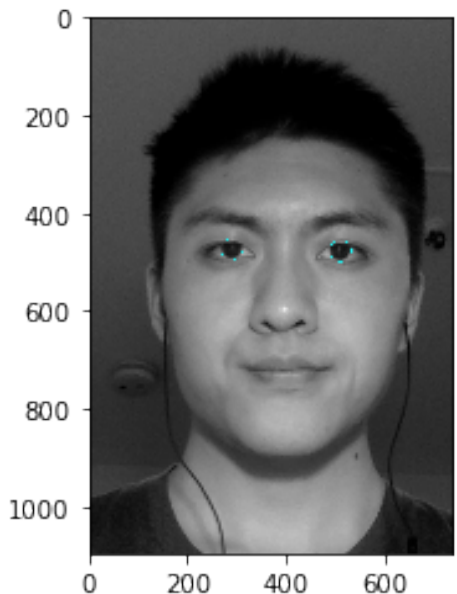

Figure 3. Example of Hough circle detection.

### 3.2.5 Skin Color Model

Using skin tone to identify faces is an old but effective technique [12]. Arguably the simplest face detection method, it relies on the fact that almost all human skin tones lie in a particular range of the $C_b$, $C_r$ color spectrum in the $YC_bC_r$ model. Different skin tones are noisy only in the $Y$ component. Thus skin is fairly easy to detect, and then simple heuristics can often identify the face and facial features with high accuracy even with no training data in nearly any light conditions. While excellent at detecting a face, the skin tone technique suffers from several key drawbacks: it does not handle occlusion well, and it can provide only a rough estimate of facial orientation. Furthermore, while not computationally expensive, it provides relatively little useful information per unit computation than our other techniques.

We chose to implement the skin color model for two principle reasons. First, it requires no training data; human skin color is well known and varies little with light conditions. Second, this model is surprisingly effective at filtering out non-face items that are mathematically similar to faces–false positives that techniques like LLR often produce. We therefore believed that the skin model will be most useful as a "first detection" algorithm, capable of quickly picking a face out of the background and a rough orientation with a complete lack of training data on a given user. That information can then be persisted into the other algorithms for finer-grained detection and tracking with few false positives.

We therefore implemented the skin color model using the approach described in Yang's seminal work [12], with approximate $Y$, $C_b$, $C_r$ bounds of:

$$80 \leq Y \leq 240$$
$$105 \leq C_b \leq 135$$
$$135 \leq C_r \leq 165$$

The technique we developed to detect a face is:

- Create a binary 2-d matrix with 1 representing pixels that fall within the skin tone range, 0 otherwise
- Fill in individual 0 pixels that have at least $n$ neighbors that are 1 to smooth out pixel noise
- Use a sliding window to identify pixel clusters of approximately the size of a face
- Rate potential faces by characteristics such as size, shape, orientation, and location on the photo
- Return the bounds of the most likely cluster as the face

If a face is detected within the image, we then proceeded to identify any eyes within the image using essentially the reverse technique by finding clusters of 0 pixels, as the eyes are almost never skin colored. Thus with some geometry and filtering similar to face detection the eyes can typically be found. The original image is then updated with bounding boxes for the face and eyes, and a green screen effect is used to easily show which pixels are not in the detected face.

We have found that this simple approach is nearly 100% effective in detecting faces with no training data. Figure 3 shows the progress we were able to make using only skin tone, no geometric help, in light of a very difficult image with numerous objects that are very visually similar to skin. We are able to detect skin fairly well, but rejection of similar non-skin needs work. E.g., the pale background is properly rejected, but the orange dog hair requires additional work. We considered using a statistical model to determine the range of color likely to be actual skin, but as discussed in the results section we soon abandoned this method because it is too slow.

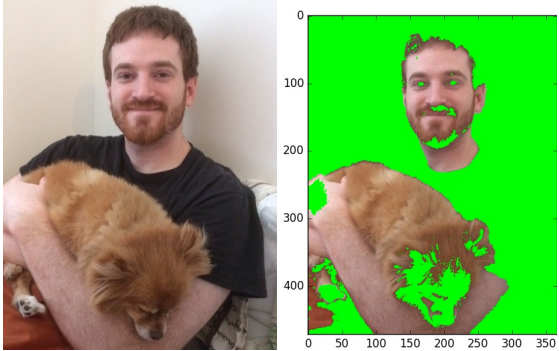

Figure 4. Skin Color Model Detection, No Filtering

### 3.2.6 Eigenfaces Model

Eigenfaces is an approach that uses principal component analysis to essentially recognize faces by decomposing a picture and using its eigenvectors to match against a set of training data. The model assumes that any human face can be expressed as a combination of a standard set of sample faces. PCA can thus reveal specific contribution amounts from a set of training samples and thus detect the makeup of a new face. [14]

The value we see in utilizing the eigenface technique is that it can recognize a new face based on a relatively limited set of training data we can package on a user's device and is known to be extremely fast. The technique is not, however, very robust to changes in expression or orientation, so we start on the assumption that users interacting with a device tend to have fairly neutral, non-changing facial expressions and orientation does not change significantly. We chose this model for the same reasons we chose the skin tone model; it is likely to be successful quickly to identify a baseline or face location, data which can then be passed to the rest of the pipeline and used in more accurate but slower techniques. Eigenfaces utilizes more assumptions (pose, expression, more training data), but if it can find a match fairly quickly, then the eigenfaces method will generally produce more accurate results than the skin color model. After initial recognition and calibration, the implementation would then "hand off" more nuanced recognition and tracking tasks to the other models discussed in this paper. Ultimately, the training data assumption proved too much for this model and we abandoned it as discussed in the results section.

### 3.2.7 Haar Cascades

Viola and Jones' seminal work on face detection used Haar features in a cascade to detect faces with high accuracy [16]. Haar features are a specific type of image feature that sums image intensities within a block and its surrounding rectangular blocks and takes the difference of these sums as a feature value. We did not implement this technique ourselves but instead relied on the OpenCV implementation; we used it because it performed substantially faster than any of our handwritten face detection algorithms but was not more accurate in our testing. Haar cascades allowed for fast localization of general areas of interest. Location of the eyes and positioning of the face were extracted and fed into our other algorithms to reduce the search area and speed up the other algorithms enough to perform in real time.

### 3.2.8 Edge/Corner Detection with Gaussian Smoothing

Implementation of the Harris corner detector [4] was necessary to extract information about the corners of the eyes. The equation, represented by

$$E(u,v) = \sigma_{x,y} w(x,y)[I(x+u, y+v) - I(x,y)]^2$$

where E represents the difference between the original and the moved window, u and v represent displacement, and I represents intensity of the window about specific coordinates. The implemented Harris corner detector seeks to maximize the variation between windows that are moved around, which is often able to help determine likely locations for the corner of the eyes.

In order to optimize the Harris corner detector to be more robust to noise, the input was often pre-processed by applying a Gaussian smoothing filter and a Canny Edge detector. The Gaussian smoothing filter, represented by the equation:

$$G(x,y) = 1/(2\pi\sigma)^2 e^{-(x^2+y^2)/(2\sigma^2)}$$

helps to reduce discolored pixels, blemishes, and general noise captured by the cameras.

### 3.3. Gaze Estimation, Measurement, and Accuracy

One proposed approach combining collected features of eye location and face orientation [7] after the application of a neural net resulted highly accurate estimates for gaze detection. In context for this project, however, approaches using neural nets are deemed infeasible due to the undesirable amount of data required to train them. As a result, we compute the estimated 3D coordinates by modeling the 3D location of the user's head and extending the line formed by the estimated gaze onto the monitoring plane. This approach may present variability due to the size of the individual's head and other factors inherent to variability in physical size, but training on the dataset to determine the features of interested per individual offers opportunities to tailor estimation per individual.

This straightforward technique makes measurement and accuracy simple; our estimation technique produces (x,y) gaze estimation coordinates on the image plane, and these can be compared with the actual values the user targets. In practice, we test our accuracy using a series of predefined points, asking the user to stare at those points, and then recording our pipeline's estimation of the user's gaze. A simple $L_2$ norm is then used to calculate the distance between estimated and actual gaze points, reported in pixels. This approach is typical of similar peer work when assessing gaze estimation accuracy.

## 4. Experiment and Results

We successfully created two working gaze estimation pipelines, although only one has reasonable accuracy. We spent most of our refining and development effort on the first pipeline. A number of the previously mentioned techniques, while promising at first, were not suitable for any pipeline.

### 4.1. Pipeline 1: Augmented Hough Circle Estimation with Edge Detectors

This pipeline was able to achieve real time gaze estimation with relatively high accuracy and reasonable latency, providing approximate gaze estimates within 1 second. The combination of components helped to minimize projection noise while running in a reasonable timeframe.

#### 4.1.1 Pipeline Components

Haar cascades: Used to quickly identify the general region of the eyes and center of the face. Other pipeline components were ran exclusively on the identified regions for the eyes and face in order to decrease runtime.

Hough transform: Applied to estimate the location of the pupils. After applying Gaussian smoothing, an edge detector, and considering the darkness of the identified pupil region to help remove noise.

Corner Detector: Used to determine relative positioning of pupils to corners of eyes for gaze estimation. The corner detector is applied to the left and right region of the eye, and the identified corners are used to select the true corners of the eyes. In for results recorded in this paper, a Harris corner detector was used. Other detectors, such as Shi and Tomasi's "Good Feature to Track" detector [11], were tested and performed without significant errors as well.

Weighted decay function: Applied in order to prevent inherent jumpiness from video recordings to severely impacting the estimated gaze. That is, given the estimated gaze coordinate from a single frame, $E_{current}$, we calculate the new estimated gaze at time t+1 as:

$$E_{t+1} = E_{current} \times \alpha + E_t \times (1 - \alpha)$$

where $\alpha$ is a weighted decay factor. In practice, a weighted decay factor of 0.1 was able to sufficiently minimize jumpiness.

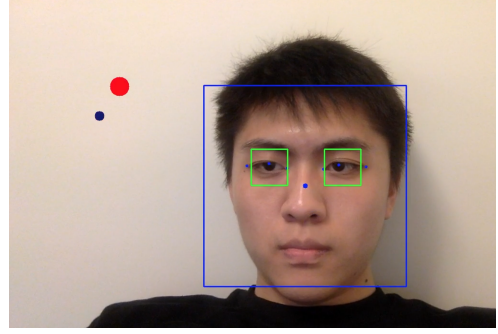

Figure 5. Pipeline with estimated (red) and actual (dark blue) gaze.

#### 4.1.2 Accuracy and Comparison

Testing different pixel coordinates on the screen, we computed the euclidean norm in pixels between the estimated gaze (Est) and the actual gaze (Act) over multiple runs.

$$Error = \sqrt{(Act_x - Est_x)^2 + (Act_y - Est_y)^2}$$

On average, the pipeline had a pixel estimation error of 90 pixels on a 13-inch Macbook pro with 1280x720 pixel screen resolution. The following numbers represent several hundred data points across three subjects. The data for different tested pixel coordinates are as follows:

| Actual X | Actual Y | Average l2 norm |
|---|---|---|
| 320.0 | 180.0 | 189.6 |
| 320.0 | 360.0 | 81.755 |
| 320.0 | 540.0 | 76.7 |
| 640.0 | 180.0 | 50.695 |
| 640.0 | 360.0 | 139.25 |
| 640.0 | 540.0 | 122.08 |
| 960.0 | 180.0 | 55.6 |
| 960.0 | 360.0 | 35.46 |
| 960.0 | 540.0 | 31.98 |

More interesting, however, is the error in millimeters divided by sections of the screen. The following results show average error and standard deviation in each of nine grid sections of the screen, subdividing the screen into 3x3 equal subparts:

Mean Error Per Subsection (mm)

|  | Left | Middle | Right |
|---|---|---|---|
| Top | $\mu = 12.3$ | $\mu = 19.7$ | $\mu = 4.6$ |
| Middle | $\mu = 12.9$ | $\mu = 22.4$ | $\mu = 5.7$ |
| Bottom | $\mu = 30.6$ | $\mu = 8.2$ | $\mu = 8.9$ |

Standard Deviation of Error Per Subsection (mm)

|  | Left | Middle | Right |
|---|---|---|---|
| Top | $\sigma = 2.0$ | $\sigma = 3.7$ | $\sigma = 2.5$ |
| Middle | $\sigma = 4.8$ | $\sigma = 3.5$ | $\sigma = 1.6$ |
| Bottom | $\sigma = 8.1$ | $\sigma = 3.6$ | $\sigma = 1.6$ |

Both the mean and standard deviation are substantially lower on the right side of the screen, but otherwise shows little pattern. Overall, standard deviation is also best on the right but more uniform than mean across the screen. This result is actually encouraging, because it implies that our precision is fairly good–the pipeline generally produces replicable results. However, some accuracy clearly varies much more. This gives us hope that with additional refinement or better calibration we could produce much more accurate results, and that the poor results in some sections are due to calibration issues rather than limitations of the technique.

The degree of accuracy achieved is usually within a few degrees of the true location. At approximately 620mm distant, our approximate average error is 13mm with average standard deviation of only about 3mm. This level of accuracy is than sufficient for medium-grained control of a device, allowing for clicking icons, panning a monitor, selecting radio buttons on most forms, and many such possible and common applications of gaze tracking. Importantly, we achieve that level of accuracy with only minimal calibration per user and only basic appearance-based techniques!

Our results compare favorably with previous work. Our average error is 10-15% lower than than other appearance-based, real-time gaze detectors such as [10], [2], and even has half the error as some simpler neural net gaze trackers [15]. As expected, our error is nearly double that of a modern convolutional neural net [18]. However, just a few sections of the screen have extremely high error (22 and 30mm), while most sections are 5-12mm. With a little work refining our techniques and focusing on removing error in those noisy sections, we are confident that we could reduce our average error to 10mm or possibly less. We hope to continue this work further to prove its potential.

#### 4.1.3 Demonstration

*Please see the supplementary materials section for links to a working demonstration and our code.*

### 4.2. Pipeline 2: OCEM and Five Point Calibration

This pipeline is comprised of 3 factors: edge detection, retina estimation, gaze estimation. While we were mildly successful in producing accurate retina estimation, the level of accuracy was not as good as pipeline 1 and the time per frame was typically 5-10 seconds.

#### 4.2.1 Pipeline Components

Edge detection: In this pipeline we use Sobel edge detection convolving in one direction to highlight the vertical aspect of the limbus and smoothed the image using a second order Gaussian function. We found that Harris and Canny edge detection did a slightly better job but Sobel was good enough and slightly computationally expensive.

Refined LLS for OCEM: We then use LLS to predict the initial iris location by fitting a circle using least squares to the edge points that we suspect to be the limbus. Then we apply a standard non-maximum noise suppression on the area where the original limbus was estimated then re-estimate the iris location within a smaller window containing each eye. By preforming LLS individually in each eye region using the original LLS as a basis for the eye regions and suppressing noise we get the best results. Then we use OCEM to refine our iris estimation.

$$\text{threshold cutoff} = (\text{max intensity} - \text{min intensity}) * 20\%$$

Accept value as above threshold if:

$$\text{pixel intensity} > \text{max intensity} - \text{threshold cutoff}$$

Calibration and gaze estimation: After defining calibration points of known location, we pair them to the estimated location of our retinas while location at various calibration locations during a calibration step. Based on these calibration-retina pairs we discern a screen to retina movement ratio. From this we can determine gaze from any estimation of our retinas by comparing the estimation to their resting location (center of the screen and first calibration point).

$$\text{screen ratio} = \frac{\text{average} \Delta screen}{\text{average} \Delta retina}$$

$$\hat{gaze} = \text{screen ratio} * \Delta retina$$

#### 4.2.2 Accuracy and Comparison

The error of this pipeline was typically greater than 150 pixels, or nearly 30mm, substantially worse than pipeline 1, and thus we focused our remaining development time on improving pipeline 1. A greater than 30mm error does not compare favorably with existing techniques, but we do think this pipeline's techniques have potential if refined.

### 4.3. Alternative Methods

Some of our techniques were found to be unsuitable for pipeline use either because they were too slow or too inaccurate to contribute substantially.

**Skin Tone Model** The skin tone model proved excellent at identifying faces and eyes, but its resolution was not sufficient to accurately locate the pupils. The pupils were simply not distinct enough under normal light conditions and the eye too irregular to develop a coherent model based on color appearance alone. Furthermore, processing a single frame using the skin tone model typically required 1-3 seconds at very low resolution, and 10+ seconds at a resolution sufficiently high to identify the pupils with reasonable accuracy. Thus the technique proved both too inaccurate and too slow to contribute substantially to any of our pipelines.

Embedding geometric techniques, eg LLS, to refine the rough eye location produced by the skin tone model may have proven accurate enough to use, but this addition would only increase processing time. Given the success of other techniques we halted development on skin tone.

The basic technique does work, however, and given a tenfold increase in speed may have proven useful. Figure 6 shows some examples of the model in action.

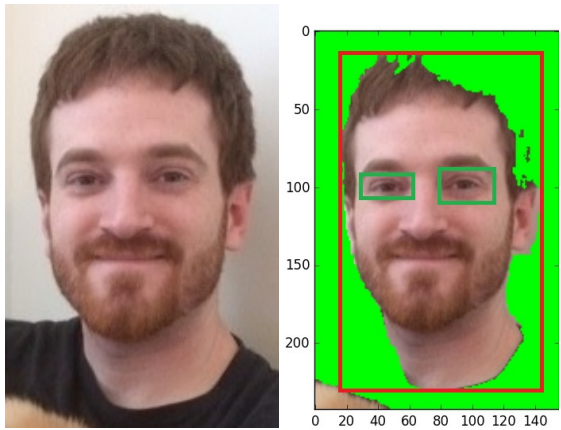

Figure 6. Skin Color Model, Original and Face/Eye Detection

**Eigenfaces Model** Although promising, the eigenfaces model proved to require too much training data to operate successfully. We built a basic pipleline to generate the face template needed for recognition, but using only one or two faces resulted in a template that was too indistinct to meaningfully match most faces. An example of the templating run is shown in Figure 7 on left, while typical templates are shown on the right. You can see that using a single subject in highly similar poses does not generalize the picture well, and rotations about the head are fairly obvious rather than creating a nice "average" image for templating.

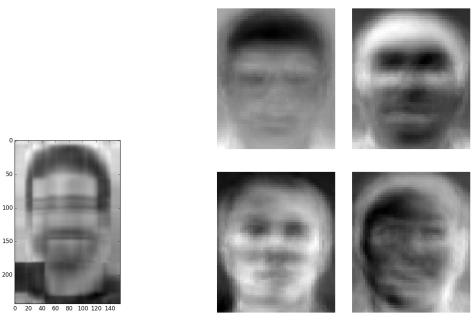

Figure 7. L: Our Template, R: Typical Templates

*Typical templates drawn from AT&T Labs [21].*

We abandoned the eigenfaces technique after it became clear that we could not generate enough training data from a single person to effectively use this technique, and the pose and expression variance of the eigenfaces technique further hampered its usefulness. However, the technique itself is sound, and if a "perfect" training set were created from carefully chosen example faces, it may yet prove useful and effective on minimal training data. We chose to invest our effort in the other pipelines rather than try to curate the perfect small training set.

## 5. Conclusion

Our experiments show that it is possible to build a highly accurate gaze tracker using just simple face and eye recognition techniques that are computationally feasible on modern mobile devices. Although our accuracy does not match that of a neural network, it is substantially faster and easier to implement. Our pipeline is able to correctly estimate gaze with an average error of only 13mm – less than 5% of typical screen height. Furthermore, our variance is typically very small, only 2-3mm. Our pipeline works with only a short calibration step required of each user. The speed of our pipeline and its minimization of training data make it suitable for further research and, eventually, ease of consumer adoption.

We also discovered that a number of techniques, while effective at localizing the face and eyes, are not suitable to the gaze estimation task. These techniques, such as LLS and the skin tone method, may yet prove useful with further enhancements. Skin tone in particular showed promising results but was simply too slow; perhaps in 10 years it will be feasible to run this technique on mobile devices. In the meantime, the other techniques are sufficient to estimate gaze reasonably for medium-grained tasks such as paging, icon selection, monitor selection, or other such tasks.

Traditionally our eyes have been limited to conveying emotion or context clues to those we are speaking to or looking at, however they will soon be a source of tangible output. Gaze estimation is an important and growing segment of human-computer interaction research, and the next decade should see the widespread adoption of gaze as a fundamental input method with many applications.

## References

[1] A. R. A. George. Real-time eye gaze direction classification using convolutional neural network. https://arxiv.org/pdf/1605.05258.pdf, 2016.

[2] D. Z. A. F. A. Sippl, C. Holzmann. http://mint.fh-hagenberg.at/wp-content/uploads/2014/07/ami2010_sippl.pdf, 2010.

[3] M. M. C. Morimoto. Eye gaze tracking techniques for interactive applications. https://www.researchgate.net/publication/222135492_Eye_gaze_tracking_techniques_for_interactive_applications, 2005.

[4] M. S. Chris Harris. A combined corner and edge detector. http://www.bmva.org/bmvc/1988/avc-88-023.pdf, 1988.

[5] E. W. et al. Eyecontact: Scleral coil eye tracking for virtual reality. http://www.cs.cmu.edu/~ltrutoiu/pdfs/ISWC_2016_trutoiu.pdf, 2016.

[6] H. K. et al. An iris detection method using the hough transform and its evaluation for facial and eye movement. http://staff.itee.uq.edu.au/lovell/aprs/accv2002/accv2002_proceedings/Kashima877.pdf, 2012.

[7] K. K. et al. Eye tracking for everyone. https://people.csail.mit.edu/khosla/papers/cvpr2016_Khosla.pdf, 2016.

[8] C. S. H. Crane. Accurate three-dimensional eye-tracker. https://www.osapublishing.org/ao/abstract.cfm?uri=ao-17-5-691, 1978.

[9] E. B. Huey. The psychology and pedagogy of reading, 1908. The Macmillan Company.

[10] J. G. J. Orozco, F. Roca. https://pdfs.semanticscholar.org/5500/9192a9a04eaf7ad6eccae862aff70cccba75.pdf, 2015.

[11] C. T. J. Shi. Good features to track. Proceedings of the IEEE Conference on Computer Vision and Pattern Recognition, pages 593-600, 1994.

[12] A. W. Jie Yang. A real-time face tracker. http://isl.anthropomatik.kit.edu/pdf/Yang1996.pdf, 1996.

[13] R. S. R. Kyung-Nam Kim. Vision-based eye-gaze tracking for human computer interface. http://www.umiacs.umd.edu/users/knkim/paper/ieee_smc99_hci-eyegaze.pdf, 1999.

[14] M. K. L. Sirovich. Low-dimensional procedure for the characterization of human faces. http://engr.case.edu/merat_francis/EECS%20490%20F04/References/Face%20Recognition/LD%20Face%20analysis.pdf, 1987.

[15] A. J. N. Piratla. http://dl.acm.org/citation.cfm?id=773087, 2002.

[16] M. J. P. Viola. Robust real-time face detection. http://www.vision.caltech.edu/html-files/EE148-2005-Spring/pprs/viola04ijcv.pdf, 2003.

[17] J. F. Philip Wilson. Facial feature detection using haar classifiers. http://dems3d.hebergratuit.net/pdf/Facial%20feature%20detection%20using%20Haar.pdf?i=1, 2006.

[18] P. V. R. Konrad, S. Shrestha. https://web.stanford.edu/class/cs231a/prev_projects_2016/eye-display-gaze-2.pdf, 2016.

[19] H. C. T. Cornsweet. Accurate two-dimensional eye tracker using first and fourth purkinje images. https://www.osapublishing.org/josa/abstract.cfm?uri=josa-63-8-921, 1973.

[20] S. S. Vijayalaxmi, P.Sudhakara Rao. Neural network approach for eye detection. https://arxiv.org/pdf/1205.5097.pdf, 2016.

[21] Wikipedia. Eigenface. https://en.wikipedia.org/wiki/Eigenface.

**Supplementary Material**

**Long Demonstration of Pipeline:**
Video 1: https://www.youtube.com/watch?v=ZRQHb7BfscM
Video 2: https://www.youtube.com/watch?edit=vd&v=37o1vHAarBE
*Please see the last minute of our presentation for a voiceover description.*

**Code:** https://github.com/Georgehe4/cs231a-project
*Each folder of the code should have a README of how to run the enclosed pipeline; pipeline 1 is recommended for review. The exception is alternative methods, which were abandoned, although they do work, but would require some tuning to exactly recreate the images in this paper (or reversion to an older commit).*

**Presentation:** https://youtu.be/7JWhXVbzuCI